\documentclass[journal,10pt]{IEEEtran}
\usepackage[T1]{fontenc}
\usepackage[utf8]{inputenc}
\usepackage{amsmath,amssymb}
\usepackage{graphicx}
\usepackage{booktabs,array,multirow}
\usepackage{cite}
\usepackage{stfloats}
\usepackage{microtype}
\usepackage[hidelinks]{hyperref}
\usepackage{tikz}
\usetikzlibrary{arrows.meta,positioning,fit,calc,shapes.geometric}
\usepackage{balance}
\hypersetup{pdfauthor={Lijun Yang, Jian Zhang, Manrong Liu},pdftitle={EEG-Centered Multimodal Emotion Recognition With Multiscale Feature Learning and Cross-Modal Representation Fusion}}

\title{Multiscale Temporal–Spectral–Spatial EEG Learning With Modality-Specific and Modality-Invariant Representations for Emotion Recognition}
\author{Jian Zhang, Manrong Liu, Yilei Huang, Wenhua Liu, Xiaohui Yang and Lijun Yang* %
\thanks{Jian Zhang, Manrong Liu ,Xiaohui Yang and  Lijun Yang are with the Henan Engineering Research Center for Artificial Intelligence Theory and Algorithms, School of Mathematics and Statistics, Henan University, Kaifeng, 475004, China.}
\thanks{Yilei Huang and Wenhua Liu are with The Second Clinical College, Henan University, Kaifeng, 475004, China.}
\thanks{*Corresponding author: Lijun Yang (yanglijun@henu.edu.cn).}}
\begin{document}
\maketitle

\begin{abstract}
Multimodal physiological signals provide complementary cues for emotion recognition. They also raise two related challenges. Emotion-related EEG patterns vary across time, frequency bands, electrodes, and brain regions. Different physiological modalities also contain both modality-specific and shared affective information. This paper presents GMSI-Net, an EEG-centered multimodal emotion recognition framework. It has three key components. The temporal-spatial-spectral-region (TSSR) encoder models multiscale temporal-channel patterns, differential entropy features, and local-to-global brain-region relations. The modality-specific representation (MSR) branch uses bidirectional cross-modal attention and learnable gates. It controls information transfer while retaining modality-specific cues. The modality-invariant representation (MIR) branch learns a shared affective representation. An adaptively weighted contrastive loss aligns this representation across modalities. Independent multiscale Conv-Transformer encoders process the peripheral physiological signals. GMSI-Net is evaluated on DEAP and MAHNOB-HCI using within-subject random ten-fold cross-validation. The model reaches $96.21\pm1.47\%$ and $96.41\pm1.60\%$ for valence and arousal on DEAP. It reaches $94.10\pm2.86\%$ and $94.48\pm3.77\%$ on MAHNOB-HCI. These results indicate that combining EEG features with complementary physiological information can improve emotion recognition.
\end{abstract}

\begin{IEEEkeywords}
multimodal emotion recognition,graph neural network, modality-specific representation, modality-invariant representation.
\end{IEEEkeywords}

\section{Introduction}
Affective computing concerns the recognition and interpretation of human affective states by computational systems \cite{picard,russell}. Emotion recognition is used in adaptive human--computer interaction, assistive technology, and affect-aware interfaces. Facial expression and speech are informative, although their measurement can be affected by lighting, background noise, and the extent to which emotion is outwardly expressed. Physiological signals provide another source of evidence. Electroencephalography (EEG) records neural activity at high temporal resolution. Electrooculography (EOG), electromyography (EMG), electrocardiography (ECG), and galvanic skin response (GSR) reflect ocular, muscular, cardiac, and electrodermal responses, respectively. DEAP and MAHNOB-HCI are widely used datasets for evaluating physiological emotion recognition \cite{deap,hci}.

Two forms of variation matter when working with physiological signals. An emotional response changes over time, and its EEG correlates may appear in different frequency bands or scalp regions. Peripheral responses can evolve at different rates. Feature extraction thus needs to account for differences between signals. Their integration also needs to retain information specific to individual sensors.

EEG is difficult to represent with a single type of feature. Even a short segment contains temporal fluctuations, channel-dependent activity, and frequency-band patterns. A filter with one receptive field may miss patterns at other temporal scales. Frequency descriptors summarize oscillatory activity but contain limited temporal-order information. Independent channel encoding also provides little information about spatial relations. Differential entropy (DE) is commonly used to characterize frequency-band statistics \cite{de}. Spatial dependencies can be modeled with graph neural networks. DGCNN learns dynamic relations between electrodes \cite{dgcnn}, while LGGNet and PGCN use local-to-global and hierarchical graph structures \cite{lggnet,pgcn}. Attention-based methods model longer-range dependencies \cite{hiertrans}. In many designs, however, temporal, frequency, and brain-region features are handled in separate paths. Bringing them into a single EEG representation remains a practical difficulty.

Temporal features retain the sequence of EEG changes; spectral features describe band-specific activity, and region-level features reflect spatial organization. When each is compressed independently, subsequent graph operations may no longer preserve where the temporal and spectral information originated. This motivates interactions between these features before brain-region aggregation.

Combining physiological modalities raises a different issue. EEG and peripheral recordings have different channel counts, distributions, and noise levels. Concatenation keeps the encoded features but cannot distinguish useful transfer from irrelevant content. Cross-modal attention introduces selective exchange, although the attended features may still contain noise. Learning a common space captures affective information present in several sensors, but a fully invariant representation can discard cues that occur in only one modality. Cross-modal attention, representation learning, and feature reweighting have been explored to manage this trade-off \cite{emotionmeter,mhca,fu}.

GMSI-Net uses different encoders for EEG and peripheral signals. Temporal, spectral, and brain-region operations are combined in the EEG encoder; peripheral signals are processed by separate convolutional and sequential encoders. Two paths then work on the encoded features. One exchanges information between modalities while retaining their individual features. The other learns affective information shared by synchronized signals. Treating these operations separately allows complementary and common information to be modeled without assigning them the same role.

GMSI-Net is an EEG-centered multimodal emotion recognition model. Its temporal-spatial-spectral-region (TSSR) encoder handles the EEG representation. A modality-specific representation (MSR) branch performs gated bidirectional cross-modal attention, and a modality-invariant representation (MIR) branch learns shared affective features with adaptive contrastive alignment. The contributions are as follows:
\begin{itemize}
\item TSSR combines multiscale temporal-channel attention, five-band DE features, and local-to-global brain-region graph learning. Temporal, spectral, and spatial information enters a unified EEG representation.
\item The MSR branch uses pairwise bidirectional attention with learnable gates. Features can be exchanged between modalities without removing the direct path from each modality's original representation.
\item The MIR branch introduces sample-dependent contrastive weights for shared representation learning. Its output is combined with MSR for classification. The experimental analysis covers DEAP and MAHNOB-HCI, including component ablations, modality combinations, and brain-region comparisons.
\end{itemize}

\section{Related Work}
\subsection{EEG-Based Emotion Recognition}
EEG emotion recognition has been studied using temporal, frequency-band, and spatial information. Spectral descriptors characterize changes in rhythmic activity; DE is one widely used statistic for individual frequency bands \cite{de}. Convolutional networks extract short-range temporal patterns, whereas recurrent and attention-based models can consider longer sequences. SSA-MTSNet combines spectral-spatial attention, multiscale convolution, and recurrent aggregation \cite{ssamts}. These methods indicate that temporal context and spectral content contribute different information to emotion recognition.

Graph-based methods focus on spatial relations between EEG recordings that are not readily represented by a regular image grid. DGCNN learns relations among electrodes dynamically \cite{dgcnn}. LGGNet builds local and global graph representations, and PGCN uses a hierarchical spatial structure \cite{lggnet,pgcn}. Aggregating channels by region reduces the number of graph nodes and introduces a structural prior. This also makes the quality of electrode-level features important: regional aggregation should not remove the local responses on which it depends.

An electrode graph captures relations among recording sites. A region graph instead operates on features aggregated from anatomical groups. With fewer nodes, region-level learning is more compact, but it depends on the information retained during aggregation. Local-to-global approaches connect these two levels \cite{lggnet,pgcn}.

Temporal and spatial attention, graph Transformers, and multi-view EEG learning have also been investigated \cite{hiertrans,stgt,dgat2025}. Attention weights can change with the input EEG segment, while graph operations account for broader spatial relations. Adding both to a network is not sufficient by itself. Their interaction determines whether temporal, spectral, and spatial features contribute to the same representation. TSSR links these views before producing the EEG embedding for multimodal fusion.

\subsection{Multimodal Physiological Emotion Recognition}
Central and peripheral signals respond to related affective processes, but their measurements are not equivalent. EmotionMeter combines physiological representations for multimodal affect recognition \cite{emotionmeter}. Other work has used cross-modal attention, complementary pattern learning, and feature reweighting \cite{mhca,fu,c2pci}. Cross-modal attention lets information from one sensor influence another. Whether that transfer helps depends on the source features and how closely the modalities correspond in a given observation.

Attention need not be symmetric: the information useful from modality $A$ to modality $B$ can differ from that useful in the reverse direction. A learnable gate can limit a transferred update when it is unhelpful, leaving more of the target representation intact. This differs from applying one global weight to an entire modality.

Physiological information can be fused at the feature, decision, or representation level. Feature-level methods concatenate encoded signals, while decision fusion combines predictions from separate classifiers. Representation-level fusion learns interactions before classification. Attention and gating provide ways to vary these interactions across samples. A feature transferred through cross-modal attention, however, is not necessarily shared in its statistical form. Muscular and neural responses can each predict emotion without being interchangeable.

Shared-space learning addresses information that is consistent across sensors. Contrastive objectives bring synchronized observations closer and separate unmatched samples. Such a space reflects affective factors supported by several modalities. Other details may belong to just one sensor and should remain available for prediction. Both forms of information are retained in the proposed model. MSR models gated complementary exchanges, while MIR uses adaptive contrastive alignment for shared affective feature learning.

In contrastive learning, synchronized physiological recordings provide positive pairs; unmatched observations provide negatives. Yet the relevance of a modality can vary from sample to sample. Adaptive weights allow the contribution of each modality to vary across samples.

\section{Methodology}
\begin{figure*}[tbp]
\centering
\includegraphics[width=.96\textwidth]{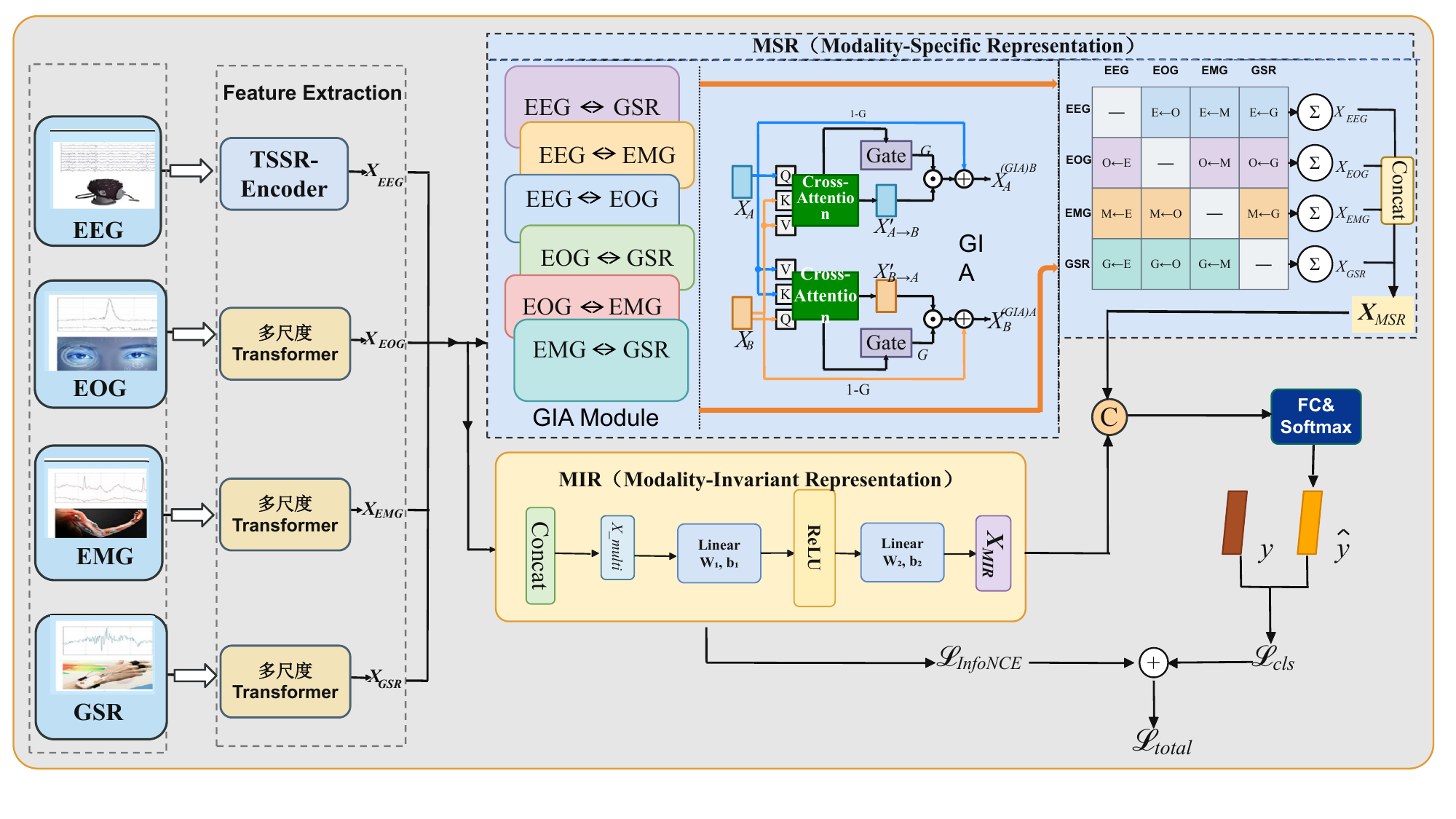}
\caption{Architecture of GMSI-Net. The EEG and peripheral encoders feed gated MSR interactions and an adaptive contrastive MIR branch. The resulting representations are fused for emotion classification.}
\label{fig:overview}
\end{figure*}

GMSI-Net uses EEG-centered feature learning followed by cross-modal representation fusion. Let $X_e\in\mathbb{R}^{C_e\times T_e}$ be an EEG segment and $X_m\in\mathbb{R}^{C_m\times T_m}$ a synchronized peripheral signal. Here, $C$ and $T$ indicate channel count and sequence length, respectively, and $M$ is the total number of modalities. The architecture is shown in Fig.~1.

The EEG path uses TSSR to extract temporal, spectral, and brain-region features. Each peripheral modality has its own multiscale Conv-Transformer encoder for temporal feature extraction. The encoded outputs are projected to a common feature width before fusion.

The fusion stage has two branches with different purposes. MSR uses gated bidirectional cross-modal attention to exchange complementary information while retaining modality-specific features. MIR maps the encoded signals to a shared affective representation and applies contrastive learning for cross-modal alignment. Their outputs, $X_{\mathrm{MSR}}$ and $X_{\mathrm{MIR}}$, are concatenated for classification.

\subsection{TSSR Encoder}
\begin{figure*}[!t]
\centering
\includegraphics[width=0.95\textwidth]{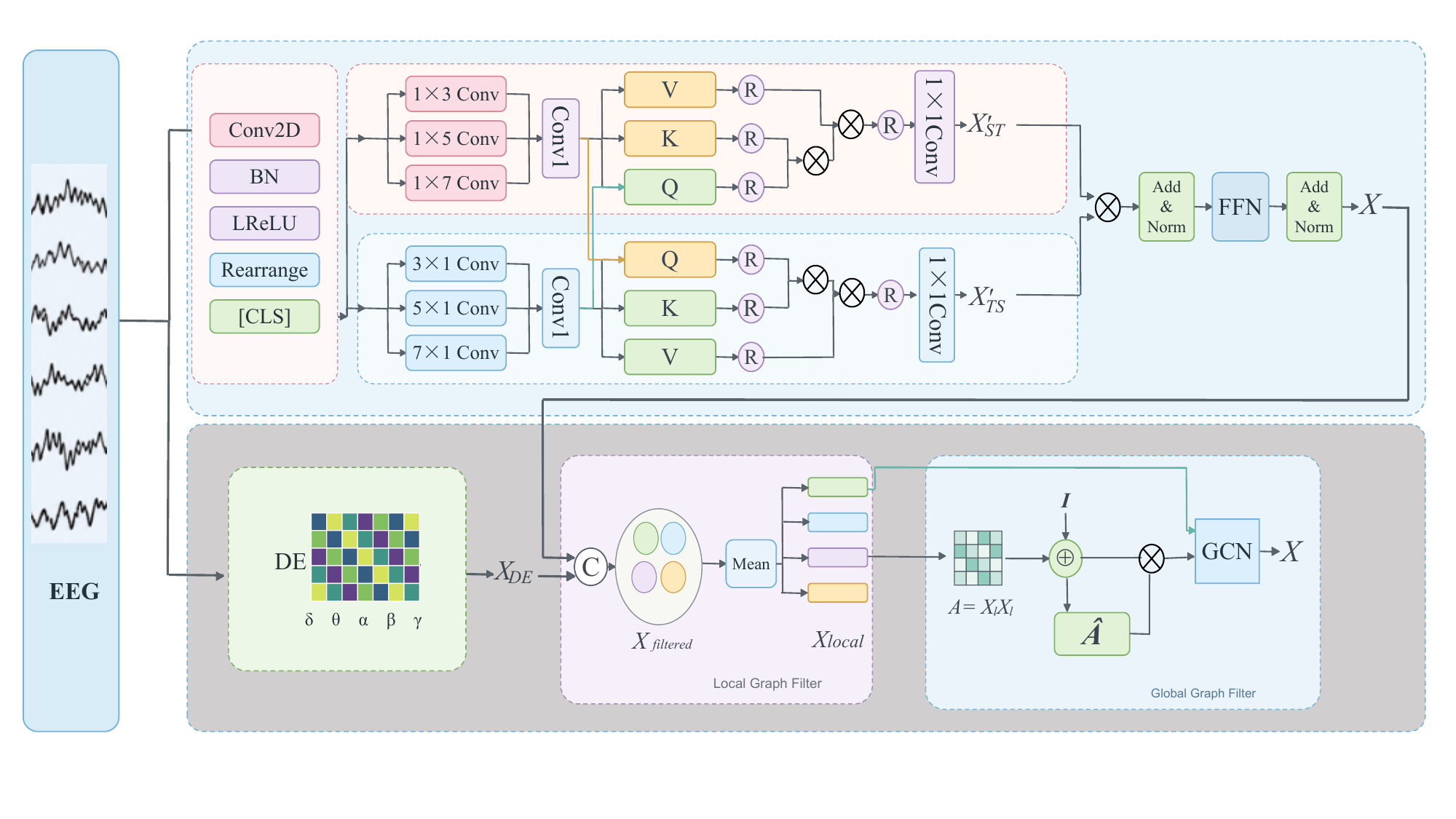}
\caption{TSSR encoder. Temporal-channel and spectral
features are fused before local and global
brain-region modeling.}
\label{fig:tssr}
\end{figure*}

TSSR combines multiscale temporal-channel features with band-specific DE and brain-region graphs. Its structure is given in Fig.~\ref{fig:tssr}.

\subsubsection{Multiscale Cross-Axis Attention}
The raw EEG segment is first embedded as a sequence. Convolution branches with different receptive fields capture local patterns at multiple scales. Cross-axis attention then relates temporal and channel features. At scale $k$, one direction is
\begin{equation}
A_{s\rightarrow t}^{(k)}=\operatorname{softmax}\left(\frac{Q_t^{(k)}K_s^{(k)\top}}{\sqrt{d}}\right)V_s^{(k)}.
\end{equation}
In this expression, $Q_t^{(k)}$ denotes the temporal query; $K_s^{(k)}$ and $V_s^{(k)}$ denote the channel-wise key and value. The superscript $k$ identifies the scale and $d$ is the attention dimension. The two attention directions are combined with scale-specific features:
\begin{equation}
Z_k=\operatorname{LN}\left(U_k^t+U_k^s+W_o[A_{s\rightarrow t}^{(k)};A_{t\rightarrow s}^{(k)}]\right).
\end{equation}
The terms $U_k^t$ and $U_k^s$ are temporal and channel features, $A_{t\rightarrow s}^{(k)}$ is the reverse attention output, and $W_o$ is a learnable projection. Brackets indicate concatenation; $\operatorname{LN}$ is layer normalization. Separate projections are used along the two axes. Residual features preserve information at each scale, and the attention outputs link temporal and electrode relations before DE and brain-region modeling.

\subsubsection{Frequency-Band Differential Entropy}
DE features are computed in the five bands $\delta$, $\theta$, $\alpha$, $\beta$, and $\gamma$. For an EEG variable approximated as Gaussian, DE is
\begin{equation}
h(X)=\frac{1}{2}\log(2\pi e\sigma^2).
\end{equation}
Here, $X$ denotes the band-limited EEG signal, $\sigma^2$ its variance, and $e$ Euler's number. The band-wise DE features join the temporal-channel representation, so both types of information are available to the graph encoder.

Band-wise DE is extracted for each channel and fused with the learned temporal-channel features. The combined representation is then passed to the graph stage.

\subsubsection{Local-to-Global Brain-Region Graph}
After temporal and spectral fusion, EEG features are grouped by brain region. Within each region, a learnable filter updates electrode-level node features:
\begin{equation}
X^{l}_{\mathrm{filtered}}=\Phi_{\mathrm{ReLU}}(W_{\mathrm{local}}\odot X^{l}_{TS}-b_{\mathrm{local}}).
\end{equation}
The input $X_{TS}^{l}$ contains fused features for region $l$. $W_{\mathrm{local}}$ and $b_{\mathrm{local}}$ are learnable filter parameters, $\odot$ means element-wise multiplication, and $\Phi_{\mathrm{ReLU}}$ applies ReLU. The filtered nodes are aggregated within each region to form $X_{\mathrm{local}}$. A graph is then constructed between regions. Using the normalized adjacency $\hat A=D^{-1/2}AD^{-1/2}$, graph convolution takes the form
\begin{equation}
X_{\mathrm{global}}=\sigma(\hat AX_{\mathrm{local}}W_g).
\end{equation}
Here, $A$ is the region adjacency matrix, $D$ its degree matrix, $W_g$ the graph-convolution weight, and $\sigma$ the activation function. Local filtering models relations within regions, while graph convolution carries information between them. The main experiments use the DL partition; Section~\ref{sec:discussion_region} compares the partition schemes.

The graph operates on the combined temporal and DE representation. Local filtering keeps electrode-level detail; global convolution models interactions among regions. A final projection matches the feature width of the peripheral encoders.

\subsection{MSR Representation Learning}
MSR transfers complementary information without discarding the original modality features. Each modality pair $(A,B)$ is processed in both directions, $A\leftarrow B$ and $B\leftarrow A$, by a gated interaction block. The attention for $A\leftarrow B$ is
\begin{equation}
\widetilde H_{a\leftarrow b}=\operatorname{softmax}\left(\frac{Q_aK_b^{\top}}{\sqrt d}\right)V_b.
\end{equation}
The query $Q_a$ comes from the target modality $a$; $K_b$ and $V_b$ are the key and value of source modality $b$, and $d$ is the attention dimension. A learnable gate controls the transferred information:
\begin{equation}
g_{a\leftarrow b}=\operatorname{sigmoid}\left(W_g\operatorname{LN}(\operatorname{AvgPool}(\widetilde H_{a\leftarrow b}))+b_g\right).
\end{equation}
Here, $\operatorname{AvgPool}$ is average pooling, $W_g$ and $b_g$ are the gate parameters for this branch, and $\operatorname{sigmoid}$ gives the gate values. The resulting feature is
\begin{equation}
S_{a\leftarrow b}=g_{a\leftarrow b}\odot\widetilde H_{a\leftarrow b}+(1-g_{a\leftarrow b})\odot H_a.
\end{equation}
The original target feature is $H_a$, $g_{a\leftarrow b}$ controls its update, and $S_{a\leftarrow b}$ is the gated output. With $M$ modalities there are $M(M-1)/2$ bidirectional interaction blocks. Outputs directed to the same target are aggregated into its MSR vector. The residual path retains the target feature when a cross-modal update is less useful.

In each directed interaction, the receiving modality supplies the query and retains its original feature through the residual term. The source modality supplies the attended content. Reversing the interaction exchanges these roles. With several peripheral sensors, a target may receive multiple updates, each through a separate gated block. The updates are aggregated into its MSR representation before classification, so attention and residual information remain available within the same branch.

\subsection{MIR and Joint Optimization}
MIR focuses on affective information shared across modalities, complementing the modality-specific features retained by MSR. The encoded sequence $H_m$ of modality $m$ is first pooled:
\begin{equation}
h_m=\operatorname{mean}(H_m).
\end{equation}
The operator $\operatorname{mean}$ averages over the temporal dimension, giving the modality vector $h_m$. Vectors from all $M$ modalities are then concatenated and projected to the MIR representation:
\begin{equation}
X_{\mathrm{MIR}}=W_2\operatorname{ReLU}(W_1[h_1;\ldots;h_M]+b_1)+b_2.
\end{equation}
The vectors $h_1,\ldots,h_M$ cover EEG and peripheral features; $W_1,W_2$ and $b_1,b_2$ are learnable projection matrices and biases. For sample $i$, $X_{\mathrm{MIR}}^{(i)}$ is paired with $h_m^{(i)}$ as a positive example. Other samples in the batch serve as negatives. With $L_2$-normalized features, similarity is defined as
\begin{equation}
s_{ij}^{(m)}=\frac{\bar X_{\mathrm{MIR}}^{(i)\top}\bar h_m^{(j)}}{\tau}.
\end{equation}
The bars indicate $L_2$ normalization, $i$ and $j$ index samples, and $\tau$ is the temperature. Given modality scores $w_m^{(i)}$, the normalized weights are
\begin{equation}
\widetilde w_m^{(i)}=\frac{\exp(w_m^{(i)})}{\sum_{r=1}^{M}\exp(w_r^{(i)})}.
\end{equation}
The score $w_m^{(i)}$ is unnormalized, while $\widetilde w_m^{(i)}$ is its normalized weight for sample $i$. The adaptive MIR loss, shown as the InfoNCE term in Fig.~\ref{fig:overview}, is
\begin{equation}
\mathcal L_{\mathrm{MIR}}=-\frac{1}{B}\sum_{i=1}^{B}\sum_{m=1}^{M}\widetilde w_m^{(i)}\log\frac{\exp(s_{ii}^{(m)})}{\sum_{j=1}^{B}\exp(s_{ij}^{(m)})}.
\end{equation}
The batch size is $B$. The term $s_{ii}^{(m)}$ measures a positive pair, whereas $s_{ij}^{(m)}$ for $j\ne i$ refers to a negative pair. The loss encourages agreement with synchronized modalities and separation from unmatched observations.

For classification, the MSR and MIR vectors are concatenated:
\begin{equation}
h_{\mathrm{final}}=[X_{\mathrm{MSR}};X_{\mathrm{MIR}}].
\end{equation}
Here, $X_{\mathrm{MSR}}$ is the aggregated modality-specific representation; brackets denote concatenation. A fully connected layer followed by Softmax outputs class probabilities. The full training objective is
\begin{equation}
\mathcal L_{\mathrm{total}}=\mathcal L_{\mathrm{cls}}+\lambda\mathcal L_{\mathrm{MIR}}+\alpha\|\Theta\|_2^2,
\end{equation}
In this objective, $\mathcal L_{\mathrm{cls}}$ is cross-entropy loss, $\lambda$ controls the contrastive term, and $\alpha$ controls $L_2$ regularization. The trainable parameters are denoted by $\Theta$, with $\|\Theta\|_2^2$ their squared $L_2$ norm. Contrastive alignment is used during training.

At inference, prediction uses the encoders, MSR, MIR projection, and classifier. Negative pairs are needed for the training objective but are not constructed for prediction, so contrastive alignment does not require a separate inference stage.

\section{Experiments}
\subsection{Datasets and Preprocessing}
DEAP contains recordings from 32 participants, each of whom watched 40 one-minute music videos \cite{deap}. The inputs comprise 32 EEG channels, two EOG channels, two EMG channels, and one GSR channel. Preprocessed signals are sampled at 128 Hz. The 3-s resting segment is used for baseline correction. Each trial is then divided into non-overlapping 1-s windows. Valence and arousal labels are binarized at 5.

MAHNOB-HCI contains synchronized physiological and audiovisual recordings \cite{hci}. The experiments use 27 valid subjects and physiological inputs comprising 32-channel EEG, three ECG channels, and one GSR channel. After baseline correction and downsampling, signals are divided into non-overlapping 1-s windows. The threshold for both valence and arousal is 5, as in DEAP. Both datasets are evaluated using within-subject random ten-fold cross-validation. Accuracy is reported as the mean and standard deviation across subjects. The data split strategy is kept unchanged in comparisons between model variants.

\subsection{Experimental Setup}
Valence and arousal are trained as separate classification tasks. Training uses the joint classification and contrastive learning objective; Table~\ref{tab:settings} lists the hyperparameters for both datasets.

\begin{table}[!htbp]
\centering
\caption{Training and Reproducibility Settings}
\label{tab:settings}
\renewcommand{\arraystretch}{1.05}
\setlength{\tabcolsep}{3pt}
\begin{tabular*}{\columnwidth}{@{\extracolsep{\fill}}lcc@{}}
\toprule
Setting & DEAP and MAHNOB-HCI \\
\midrule
Brain partition & \multicolumn{2}{c}{DL} \\
Batch size & \multicolumn{2}{c}{60} \\
Learning rate & \multicolumn{2}{c}{0.002} \\
Adam $(\beta_1,\beta_2)$ & \multicolumn{2}{c}{$(0.5,0.9)$} \\
Weight decay & \multicolumn{2}{c}{0.005} \\
$\lambda,\tau$ & \multicolumn{2}{c}{$0.2,\ 0.2$} \\
Random seed & \multicolumn{2}{c}{42} \\
\bottomrule
\end{tabular*}
\end{table}

The number of training epochs was examined separately for each dataset. Accuracy reached its highest observed level at approximately 100 epochs on DEAP and 50 on MAHNOB-HCI; these values were used in the final experiments. Reported accuracies are subject-level means with standard deviations.

\subsection{Experimental Results}
On DEAP, GMSI-Net obtains $96.21\pm1.47\%$ valence accuracy and $96.41\pm1.60\%$ arousal accuracy. The MAHNOB-HCI results are $94.10\pm2.86\%$ and $94.48\pm3.77\%$, respectively. Comparisons with the selected methods are given in Tables~\ref{tab:deap_compare} and \ref{tab:hci_compare}.

\begin{table}[!htbp]
\caption{Classification Accuracy on DEAP}
\label{tab:deap_compare}
\centering\scriptsize
\setlength{\tabcolsep}{1.4pt}
\begin{tabular}{@{}>{\raggedright\arraybackslash}p{.27\columnwidth}>{\raggedright\arraybackslash}p{.36\columnwidth}>{\centering\arraybackslash}m{.16\columnwidth}>{\centering\arraybackslash}m{.16\columnwidth}@{}}
\toprule
Method & Modality & \multicolumn{2}{c}{Accuracy (\%)} \\
\cmidrule(lr){3-4}
 & & Valence & Arousal \\
\midrule
Ding et al.~\cite{mhca} & EEG+PPS & 94.88 & 95.26 \\
GBV-Net~\cite{gbv} & EEG+PPS+Face & 95.10 & 95.65 \\
Att-CNN-GRU~\cite{att1d} & EEG+ECG & 92.50 & 91.30 \\
Diff-MT~\cite{diffmt} & EEG+EOG+EMG+GSR & 94.83 & 93.95 \\
HCNN~\cite{hcnn} & EEG+Face & 87.84 & 88.09 \\
\textbf{GMSI-Net} & \textbf{EEG+EOG+EMG+GSR} & \textbf{$96.21\pm1.47$} & \textbf{$96.41\pm1.60$} \\
\bottomrule
\end{tabular}
\end{table}

Valence and arousal accuracy are close on DEAP. The results are obtained by combining the structured EEG representation with peripheral information through the two fusion branches.

\begin{table}[!htbp]
\caption{Classification Accuracy on MAHNOB-HCI}
\label{tab:hci_compare}
\centering\scriptsize
\setlength{\tabcolsep}{1pt}
\begin{tabular}{@{}>{\raggedright\arraybackslash}p{.29\columnwidth}>{\raggedright\arraybackslash}p{.32\columnwidth}>{\centering\arraybackslash}m{.19\columnwidth}>{\centering\arraybackslash}m{.19\columnwidth}@{}}
\toprule
Method & Modality & \multicolumn{2}{c}{Accuracy (\%)} \\
\cmidrule(lr){3-4}
 & & Valence & Arousal \\
\midrule
HCNN~\cite{hcnn} & EEG+Face & 91.40 & 90.72 \\
STST~\cite{stst} & EEG+micro-expression & 86.10 & 85.70 \\
DMMVAE-LSTM$^{*}$~\cite{dmmvae} & EEG+Face & $93.92\pm1.80$ & $93.18\pm1.90$ \\
DeepVADNet$^{\dagger}$~\cite{deepvad} & EEG+bio+Face & 89.98 & 88.60 \\
Liu et al.$^{\ddagger}$~\cite{liu2026} & EEG+Speech & 81.09 & 80.11 \\
\textbf{GMSI-Net} & \textbf{EEG+ECG+GSR} & \textbf{$94.10\pm2.86$} & \textbf{$94.48\pm3.77$} \\
\bottomrule
\end{tabular}
\par\vspace{3pt}\noindent\parbox{.98\columnwidth}{\fontsize{7.2}{8.4}\selectfont Note: $^{*}$No-pre-trained-VAE variant from the cited study; $^{\dagger}$dimension-wise results from authors\textquotesingle{} released table; $^{\ddagger}$cross-subject validation.}
\end{table}

The two affective dimensions yield similar results on DEAP: $96.21\%$ for valence and $96.41\%$ for arousal. On MAHNOB-HCI, accuracy is $94.10\%$ and $94.48\%$. The same feature extraction and fusion design is used for both tasks.

The DEAP comparison includes EEG-plus-physiology methods by Ding et al. and Diff-MT, alongside methods that also use visual input. Among the listed results, GMSI-Net has the highest valence and arousal accuracies. On MAHNOB-HCI, DMMVAE-LSTM has the nearest reported valence result, while the listed arousal accuracy of GMSI-Net is higher. These comparisons place the model alongside recent multimodal physiological emotion recognition methods.

Variation across subjects differs between datasets. Standard deviations are $1.47$ and $1.60$ percentage points on DEAP, versus $2.86$ and $3.77$ points on MAHNOB-HCI.

\begin{table*}[!b]
\centering
\caption{Ablation Study on DEAP and MAHNOB-HCI}
\label{tab:ablation}
\footnotesize
\renewcommand{\arraystretch}{1.15}
\setlength{\tabcolsep}{5pt}
\begin{tabular*}{\textwidth}{@{\extracolsep{\fill}}lcccc@{}}
\toprule
& \multicolumn{2}{c}{DEAP} & \multicolumn{2}{c}{MAHNOB-HCI} \\
\cmidrule(lr){2-3}\cmidrule(lr){4-5}
Setting & Valence & Arousal & Valence & Arousal \\
\midrule
w/o MCA encoder & $94.73\pm1.86$ & $94.96\pm2.21$ & $93.09\pm3.48$ & $92.81\pm5.10$ \\
w/o DE features & $93.61\pm2.11$ & $93.98\pm2.17$ & $73.24\pm5.29$ & $76.12\pm9.54$ \\
w/o brain-region graph & $94.54\pm1.83$ & $94.70\pm2.21$ & $92.50\pm3.39$ & $92.28\pm5.23$ \\
w/o peripheral Conv-Transformer & $93.33\pm3.10$ & $93.38\pm4.07$ & $92.82\pm3.75$ & $92.44\pm5.19$ \\
w/o cross-modal attention & $94.89\pm1.93$ & $95.22\pm2.09$ & $92.82\pm3.36$ & $92.72\pm4.79$ \\
w/o TSSR & $93.08\pm1.92$ & $93.36\pm2.56$ & $72.11\pm5.60$ & $75.55\pm9.96$ \\
w/o MSR & $95.97\pm1.54$ & $96.37\pm1.57$ & $92.82\pm3.48$ & $92.54\pm4.97$ \\
w/o MIR & $95.49\pm1.58$ & $95.90\pm1.82$ & $92.74\pm3.41$ & $92.91\pm4.55$ \\
w/o learned modality weighting & $95.66\pm1.47$ & $96.01\pm1.63$ & $92.94\pm3.20$ & $92.87\pm4.70$ \\
\textbf{GMSI-Net} & $\mathbf{96.21\pm1.47}$ & $\mathbf{96.41\pm1.60}$ & $\mathbf{94.10\pm2.86}$ & $\mathbf{94.48\pm3.77}$ \\
\bottomrule
\end{tabular*}
\end{table*}

\subsection{Ablation Study}
Table~\ref{tab:ablation} reports ablations on DEAP and MAHNOB-HCI. EEG subcomponents are assessed in the CMMF-Net configuration; TSSR, MSR, MIR, and modality weighting are assessed in GMSI-Net. For w/o TSSR, the EEG encoder is replaced by a vanilla Transformer.

Among the evaluated final-model variants, replacing TSSR produces the largest accuracy decrease. On DEAP, valence accuracy falls from $96.21\%$ to $93.08\%$, and arousal accuracy from $96.41\%$ to $93.36\%$ when the complete EEG encoder is replaced by a vanilla Transformer. The change is larger on MAHNOB-HCI: $94.10\%$ and $94.48\%$ fall to $72.11\%$ and $75.55\%$. The structured EEG encoder is thus particularly important in the MAHNOB-HCI experiments.

The fusion branches provide further gains beyond EEG encoding, though their effects are not the same on the two datasets. Without MIR, DEAP valence and arousal accuracy drop by $0.72$ and $0.51$ percentage points. Removing MSR produces smaller decreases on DEAP and larger ones on MAHNOB-HCI. Learned modality weighting also performs better than uniform weighting. Cross-modal interaction and shared-space alignment therefore contribute at different levels across the datasets.

Accuracy also declines when DE features, brain-region graphs, or peripheral encoders are removed. Both structured EEG features and multimodal interactions contribute to the predictions.

Without DE features, MAHNOB-HCI valence and arousal accuracy fall to $73.24\%$ and $76.12\%$. Spectral features have a substantial effect in this setting.

The representation-level ablations on DEAP remain relatively close to the full-model accuracy. Removing MSR changes valence by $0.24$ and arousal by $0.04$ points; removing MIR changes them by $0.72$ and $0.51$ points. On MAHNOB-HCI, the two branches make more similar contributions. DE features also affect the datasets differently. Their removal produces a modest decline on DEAP but a much larger one on MAHNOB-HCI, indicating a stronger dependence on spectral information in the latter experiments.

Taken together, the ablations identify structured EEG feature extraction and cross-modal representation learning as useful parts of the model. TSSR accounts for a large part of the EEG representation gain. MSR, MIR, and adaptive modality weighting bring additional improvements to multimodal fusion.

\subsection{Effect of Modality Combinations}
Tables~\ref{tab:modal_deap} and \ref{tab:modal_hci} compare the input combinations. The complete four-modal configuration gives the highest accuracy among the tested combinations on DEAP. On MAHNOB-HCI, the highest accuracy is obtained with EEG+ECG+GSR. Not every smaller subset benefits from an added sensor; a new modality may introduce redundancy or noise along with useful information.

All tested combinations include EEG, consistent with the EEG-centered design of GMSI-Net. These experiments test what peripheral signals add to EEG rather than whether they can replace it. Accuracy does not increase consistently as modalities are added. The usefulness of peripheral inputs depends partly on signal quality and redundancy between sensors.

For DEAP, the evaluated configurations combine EEG with different subsets of EOG, EMG, and GSR. The MAHNOB-HCI experiments instead combine EEG with ECG and GSR.

\begin{table}[!htbp]
\caption{Modality Combinations on DEAP}
\label{tab:modal_deap}
\centering
\scriptsize
\setlength{\tabcolsep}{2.2pt}
\begin{tabular}{lcc}
\toprule
Combination & \multicolumn{2}{c}{ACC $\pm$ Std (\%)} \\
\cmidrule(lr){2-3}
 & Valence & Arousal \\
\midrule
EEG & $89.45\pm4.65$ & $90.45\pm4.49$ \\
EEG+EOG & $91.82\pm3.18$ & $92.18\pm3.06$ \\
EEG+EMG & $92.17\pm3.73$ & $92.45\pm5.32$ \\
EEG+GSR & $88.99\pm4.81$ & $90.44\pm3.96$ \\
EEG+EOG+EMG & $93.83\pm1.70$ & $94.07\pm1.80$ \\
EEG+EOG+GSR & $92.57\pm1.85$ & $93.05\pm2.03$ \\
EEG+EMG+GSR & $92.34\pm2.05$ & $93.03\pm2.35$ \\
\textbf{EEG+EOG+}\newline\textbf{EMG+GSR} & \textbf{$96.21\pm1.47$} & \textbf{$96.41\pm1.60$} \\
\bottomrule
\end{tabular}
\end{table}

EEG alone reaches $89.45\%$ valence and $90.45\%$ arousal accuracy on DEAP. Adding either EOG or EMG improves both tasks, but adding GSR alone does not improve the EEG baseline. The EOG--EMG pair reaches $93.83\%$ and $94.07\%$. With all three peripheral signal types, accuracy increases to $96.21\%$ and $96.41\%$. Compared with EEG alone, these are gains of $6.76$ and $5.96$ percentage points in valence and arousal. The gains are more pronounced when complementary peripheral responses are used together than when a single peripheral sensor is added.

On MAHNOB-HCI, ECG and GSR provide cardiac and electrodermal measurements alongside EEG. Their combinations are reported separately for valence and arousal.

\begin{table}[!htbp]
\caption{Modality Combinations on MAHNOB-HCI}
\label{tab:modal_hci}
\centering
\scriptsize
\begin{tabular}{lcc}
\toprule
Combination & \multicolumn{2}{c}{ACC $\pm$ Std (\%)} \\
\cmidrule(lr){2-3}
 & Valence & Arousal \\
\midrule
EEG & $91.38\pm3.99$ & $91.38\pm5.27$ \\
EEG+ECG & $91.88\pm4.36$ & $92.93\pm4.53$ \\
EEG+GSR & $91.77\pm3.65$ & $92.14\pm4.83$ \\
\textbf{EEG+ECG+GSR} & \textbf{$94.10\pm2.86$} & \textbf{$94.48\pm3.77$} \\
\bottomrule
\end{tabular}
\end{table}

On MAHNOB-HCI, EEG alone obtains $91.38\%$ accuracy for both dimensions. Adding ECG raises arousal accuracy to $92.93\%$; EEG+GSR reaches $92.14\%$. With EEG+ECG+GSR, valence and arousal accuracy reach $94.10\%$ and $94.48\%$, improving on EEG alone by $2.72$ and $3.10$ percentage points. Both peripheral modalities contribute when used together. These results are consistent with the independent peripheral encoders and gated exchange in GMSI-Net, which combines information after each signal has been encoded.

\section{Discussion}
\subsection{Brain-Region Partition}

\label{sec:discussion_region}
Table~\ref{tab:partition} compares five brain-region partitions. H groups electrodes by hemisphere, and RL uses broad anatomical lobes. DL further divides each lobe into left and right regions~\cite{lggnet}. F~\cite{lggnet} and T~\cite{pgcn} refine the frontal and temporal regions, respectively. DL gives the highest accuracies among the tested partitions on both datasets. It retains lobar organization and hemispheric distinctions and is therefore used in the main experiments.

\begin{table}[!htbp]
\caption{Brain-Region Partition Analysis (Accuracy, \%)}
\label{tab:partition}
\centering
\footnotesize
\begin{tabular}{lcccc}
\toprule
Partition & \multicolumn{2}{c}{DEAP} & \multicolumn{2}{c}{MAHNOB-HCI} \\
\cmidrule(lr){2-3}\cmidrule(lr){4-5}
 & Valence & Arousal & Valence & Arousal \\
\midrule
H & 94.52 & 94.51 & 91.49 & 91.22 \\
RL & 95.16 & 95.27 & 91.63 & 91.13 \\
\textbf{DL} & \textbf{95.37} & \textbf{95.79} & \textbf{93.76} & \textbf{93.72} \\
F & 95.19 & 95.37 & 92.83 & 93.07 \\
T & 95.27 & 95.25 & 91.49 & 91.26 \\
\bottomrule
\end{tabular}
\end{table}

With the underlying fusion backbone, DL yields $95.37\%$ valence accuracy on DEAP; H, RL, F, and T yield $94.52\%$, $95.16\%$, $95.19\%$, and $95.27\%$. Its arousal accuracy is $95.79\%$, the highest among the tested partitions. On MAHNOB-HCI, its valence and arousal scores are $93.76\%$ and $93.72\%$. Hemisphere-only grouping retains left--right distinctions but offers little regional detail. RL adds coarse anatomical structure; F and T place more emphasis on selected frontal or temporal regions. DL keeps lobar distinctions while separating the hemispheres within each lobe, and yields the highest backbone accuracies in this comparison. These partition results concern the regional encoder. Overall GMSI-Net classification results are reported separately in Tables~\ref{tab:deap_compare} and \ref{tab:hci_compare}.

\subsection{Visualization}

\begin{figure}[htbp]
\centering
\includegraphics[width=\columnwidth]{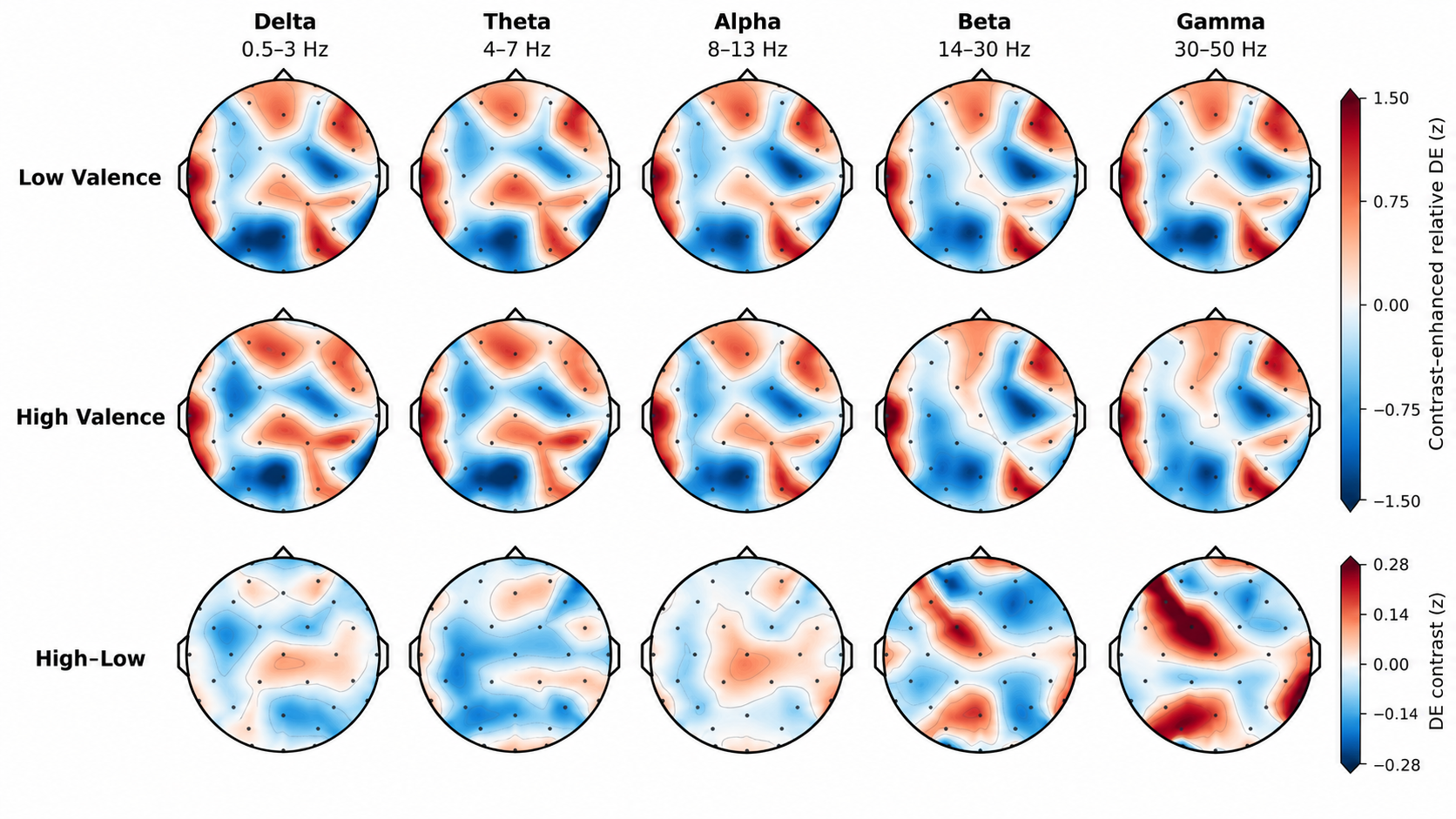}
\caption{Band-wise EEG topographies for valence on DEAP.
The first two rows show relative DE maps for low- and
high-valence states across the $\delta$, $\theta$,
$\alpha$, $\beta$, and $\gamma$ bands. The bottom row
shows the high-minus-low contrast, revealing spatial
and spectral differences between emotional states.}
\label{fig:saliency}
\end{figure}

The EEG maps in Fig.~\ref{fig:saliency} show how band-wise differential entropy (DE) varies with valence and scalp location. Low- and high-valence states display different spatial patterns in the five frequency bands. Local contrasts are most visible in the $\beta$ and $\gamma$ bands, while the differences in other bands are more limited. The changes also occur at several scalp locations. This distribution gives a data-level reason to consider both frequency-band features and spatial relations in TSSR.

\begin{figure}[htbp]
\centering
\includegraphics[width=\columnwidth]{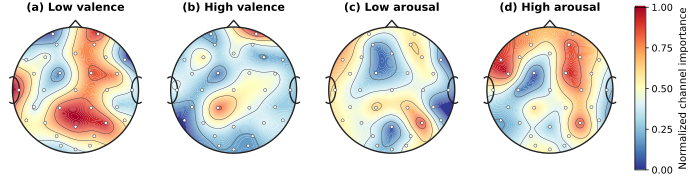}
\caption{EEG channel importance on DEAP obtained by
channel perturbation. Panels (a)--(d) represent
low valence, high valence, low arousal, and high
arousal, respectively. The normalized importance
maps illustrate task-dependent spatial contributions
of EEG electrodes.}
\label{fig:importance}
\end{figure}

Channel perturbation provides a complementary view, this time of the trained model rather than the DE maps. Fig.~\ref{fig:importance} shows that the relative importance of EEG electrodes changes between valence and arousal, and between their respective classes. The classifier shows different degrees of sensitivity to electrodes across tasks. These spatial differences are consistent with the local-to-global region modeling of TSSR, which combines electrode-level features with interactions between regions.

\begin{figure}[htbp]
\centering
\includegraphics[width=\columnwidth]{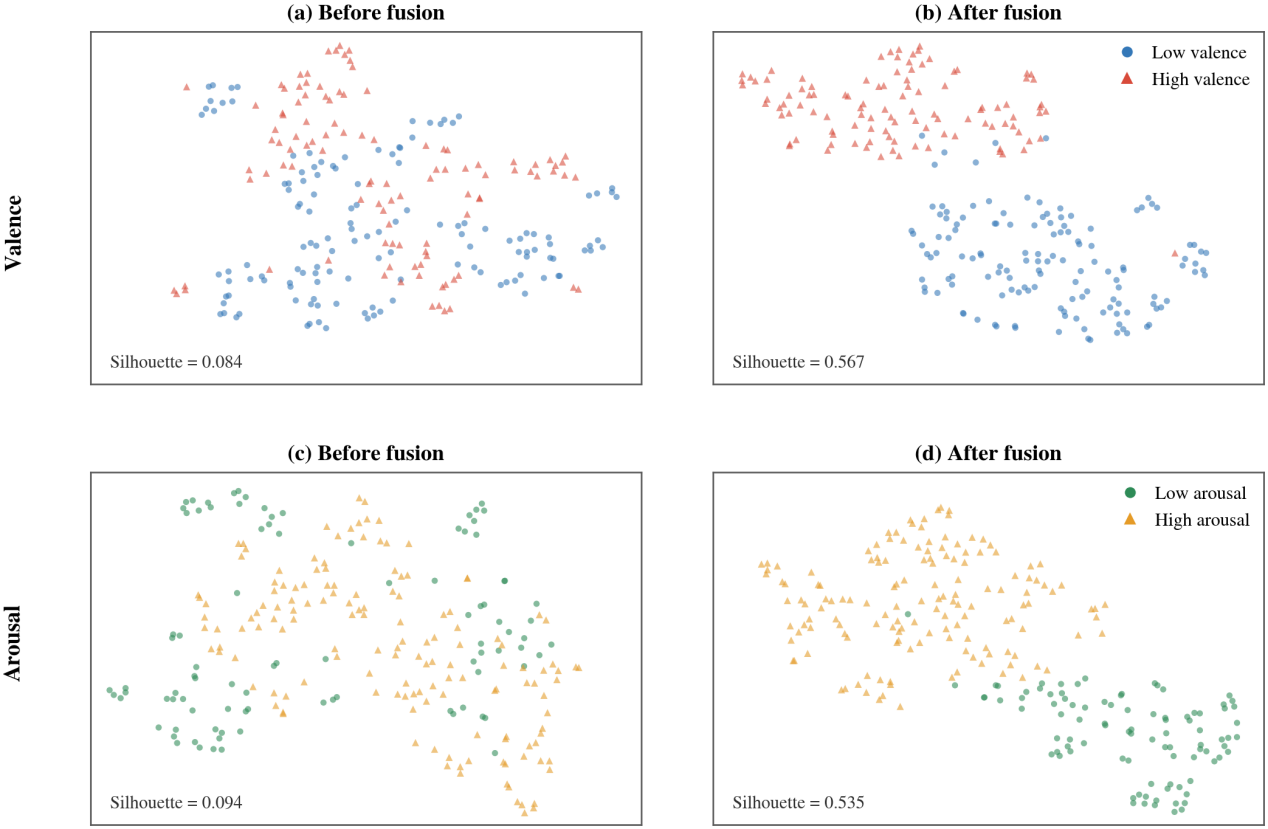}
\caption{t-SNE visualization of feature distributions
before and after GMSI-Net fusion. The top and bottom
rows correspond to valence and arousal, respectively.
The left panels show pre-fusion features, while the
right panels show fused representations. Silhouette
scores increase from 0.084 to 0.567 for valence and
from 0.094 to 0.535 for arousal.}
\label{fig:tsne}
\end{figure}

The learned representations were also examined using t-SNE. In Fig.~\ref{fig:tsne}, features from different emotional classes overlap considerably before multimodal fusion. The fused representations form more distinct clusters in the projected space. Silhouette scores increase from 0.084 to 0.567 for valence and from 0.094 to 0.535 for arousal. The projected distributions are consistent with improved class separation after MSR--MIR fusion.

\section{Conclusion}

GMSI-Net was developed for EEG-centered multimodal emotion recognition. It combines multiscale EEG feature learning with modality-specific and shared representation learning. Temporal, spectral, and spatial information is extracted from EEG, while peripheral responses provide complementary cues. The two fusion branches retain features specific to each modality and learn shared affective information.

The achieved accuracies are $96.21\pm1.47\%$ for valence and $96.41\pm1.60\%$ for arousal on DEAP, and $94.10\pm2.86\%$ and $94.48\pm3.77\%$ on MAHNOB-HCI. The experiments also show differences between components. Replacing TSSR reduces performance markedly, especially on MAHNOB-HCI. DE features have a larger effect on MAHNOB-HCI than on DEAP, and the gains from MSR and MIR differ across datasets. In the modality-combination tests, adding one peripheral sensor does not always help. Within the tested combinations, using complementary signals together yields higher accuracies. Brain-region comparisons favor structured EEG features, while the visualizations show clearer class separation after multimodal fusion.

These observations point to the value of learning EEG features across several dimensions without discarding information specific to peripheral modalities. Future work will examine cross-subject generalization, missing-modality robustness, and the efficiency of cross-modal interaction.

\section*{Acknowledgment}

The authors would like to thank the journal editor and anonymous reviewers for their constructive comments. This research was partially supported by the Key Research and Development and Promotion Special Project (Technology Development) in Henan Province (grant number 252102210067), Key scientific research projects of Henan higher education institutions (grant numbers 25A120001 and 26B320004).

\balance

\end{document}